%% file: note.tex
\documentclass[letterpaper]{article}
\usepackage[margin=1in,dvips]{geometry}
\usepackage{graphicx,psfrag,amsmath,amsthm,amssymb}
\usepackage{natbib}
\usepackage{algorithmic,algorithm}
\usepackage{url}
\usepackage{enumitem}
\usepackage{dcolumn}
\newcolumntype{d}[1]{D{.}{\cdot}{#1} }

\input{MACPdef}
\input{MACPdef-ams}

\bibpunct[, ]{(}{)}{;}{a}{,}{,} 

\usepackage{color}

\title{Projection onto the capped simplex}
\author{Weiran Wang \\
  Toyota Technological Institute at Chicago \\
  \texttt{weiranwang@ttic.edu}
  \and Canyi Lu\\
  National University of Singapore\\
  \texttt{canyilu@gmail.com}
}
\date{March 2, 2015}

\begin{document}

\maketitle

\begin{abstract}
  We provide a simple and efficient algorithm for computing the Euclidean projection of a point onto the capped simplex, formally defined as
\begin{equation*}
\min_{\mathbf{x} \in \mathbb{R}^D} \quad \frac{1}{2} {\left\lVert\mathbf{x}-\mathbf{y}\right\rVert}^2 \qquad \text{s.t.} \quad \mathbf{x}^\top \1 = s,  \quad \mathbf{0} \le \mathbf{x}\le \mathbf{1},
\end{equation*}
together with an elementary proof. Both the MATLAB and \texttt{C++} implementations of the proposed algorithm can be downloaded at \texttt{https://eng.ucmerced.edu/people/wwang5}.
\end{abstract}

\section{The Problem}
In this report, we consider the following optimization problem
\begin{subequations}
  \label{e:problem}
  \begin{align}
    \min_{\x\in\bbR^D} & \quad \frac{1}{2} \norm{\x-\y}^2 \\
    \text{s.t.} &\quad \x^\top \1 = s \\
    &\quad \0\le \x\le \1,
  \end{align}
\end{subequations}
where $s\in[0,D]$ is a parameter of the problem, $\0$ and $\1$ are vectors of $0$'s and $1$'s respectively, and $\le$ means elementwise comparison. The feasible set of this problem is the intersection of the unit cube and a hyperplane with normal $\1$. Alternatively, the feasible set is the simplex $\{\x:\ \x\ge\0,\ \x^\top \1=s\}$ with an additional capping constraints $\x\le \1$, so we call it the \emph{capped simplex}. Problem~\ref{e:problem} is a quadratic program and the objective function is strictly convex, so there is a unique solution which we denote by $\x=[x_1,\dots,x_D]^\top$ with a slight abuse of notation. 
\begin{rmk}
  This problem is a slight generalization of the projection onto the probability simplex (see \citealp{Duchi_08a,WangCarreir13a} and the references therein), which is a special case of \eqref{e:problem} by setting $s=1$ and can be solved exactly with $\calO(D\log D)$ time complexity. An elementary proof of the corresponding algorithm can be found in \citet{WangCarreir13a} and the cost mainly comes from sorting the dimensions of $\y$. Our solution to \eqref{e:problem} in this report is derived using a similar idea.
\end{rmk}
\begin{rmk}
  The constraint $\0\le \x\le \1$ in \eqref{e:problem} can be generalized to $\0\le \x\le t\1$ where $t$ is any positive number. We only need to solve the following instance of \eqref{e:problem}:\begin{align*}
    \min_{\hat{\x}}\; \frac{1}{2} \norm{\hat{\x}-\y/t}^2, \quad\text{s.t.}\quad \hat{\x}^\top \1=s/t,\quad \0\le \hat{\x} \le \1,
  \end{align*}
  and then scale its solution $\hat{\x}$ by $t$ to obtain the solution of the original problem.
\end{rmk}

\section{The algorithm}

We provide an $\calO(D^2)$ algorithm for solving \eqref{e:problem} in Algorithm~\ref{alg:proj}.
\begin{algorithm}[h!]
  \caption{Euclidean projection of a vector onto the section of cube.}
  \label{alg:proj}
  \renewcommand{\algorithmicrequire}{\textbf{Input:}}
  \renewcommand{\algorithmicensure}{\textbf{Output:}}
  \begin{algorithmic}[1]
    \REQUIRE $\y \in\bbR^D$ is sorted in ascending order: $y_1 \le y_2 \le \dots \le y_D$.
    \STATE Set $y_0=-\infty$ and $y_{D+1}=\infty$, compute partial sums $T_0=0$, and $T_k=\sum_{j=1}^{k} y_k$,  $k=1,\dots,D$.
    \FOR{$a=0,1,\dots,D$}
    \IF{$(s==D-a)$  \&\&  $(y_{a+1}-y_a \ge 1)$}
    \STATE Set b=a.
    \STATE break
    \ENDIF    
    \FOR{$b=a+1,\dots,D$}
    \STATE Compute $\gamma=\frac{s+b-n+T_a-T_b}{b-a}$.
    \IF{$(y_a+\gamma\le 0)$ \&\& $(y_{a+1}+\gamma>0)$ \&\& $(y_b+\gamma<1)$ \&\& $(y_{b+1}\ge 1)$}
    \STATE break
    \ENDIF
    \ENDFOR
    \ENDFOR
    \ENSURE $\x=[0,\dots,0,y_{a+1}+\gamma,\dots,y_b+\gamma,1,\dots,1]$.
  \end{algorithmic}\label{alg1}
\end{algorithm}

\section{The proof}
As mentioned earlier, \eqref{e:problem} has a unique solution which is characterized by its KKT system \citep{NocedalWright06a}. The Lagrangian function of the problem is
\begin{equation*}
  \calL(\x,\balpha,\bbeta,\gamma) = \frac{1}{2} \norm{\x-\y}^2 - \balpha^\top \x - \bbeta^\top (\1-\x) - \gamma(\1^\top \x-s)
\end{equation*}
where $\balpha=[\alpha_1,\dots,\alpha_D]^\top$ and $\bbeta=[\beta_1,\dots,\beta_D]^\top$ are the Lagrange multipliers for the inequality constraints $\x \ge \0$ and $\1-\x \ge \0$ respectively, and $\gamma$ is the Lagrange multiplier for the equality constraint. At the optimal solution $\x$ the following KKT conditions hold:
\begin{subequations}
  \label{e:KKT}
  \begin{align}
    x_i - y_i - \alpha_i + \beta_i -\gamma &=0, \qquad i=1,\dots,D \\
    x_i &\ge 0, \qquad i=1,\dots,D \\
    x_i &\le 1, \qquad i=1,\dots,D \\
    \alpha_i &\ge 0, \qquad i=1,\dots,D \\
    \beta_i &\ge 0, \qquad i=1,\dots,D \\
    \label{e:sum}
    \sum\nolimits_{i=1}^D x_i &= s, \\
    \label{e:comp1}
    \alpha_i x_i  &= 0 , \qquad i=1,\dots,D \\
    \label{e:comp2}
    \beta_i (1-x_i)  &= 0 , \qquad i=1,\dots,D,
  \end{align}
\end{subequations}
where \eqref{e:comp1} and \eqref{e:comp1} are complementary slackness (CS) conditions.

Without loss of generality, we assume the components of the optimal solution $\x$ are in ascending order:
\begin{align}
  0=x_1 = \dots =  x_a < x_{a+1} \le \dots \le x_b < x_{b+1} = \dots x_D=1,
\end{align}
where $a$ is the number of $0$'s in the solution while $D-b$ is the number of $1$'s in the solution. The valid ranges for $(a,b)$ are $0 \le a \le D$ and $a \le b \le D$. The KKT conditions can be simplified for different set of dimensions of the solution:
\begin{enumerate}[label=(\roman*)]
\item For $i=1,\dots,a$, the CS condition \eqref{e:comp2} indicates $\beta_i=0$, and thus
  \begin{align}\label{e:case1}
    0=x_i=y_i+\alpha_i+\gamma \ge y_i+\gamma,
  \end{align}
  where the last inequality uses the fact that $\alpha_i\ge 0$.
\item For $j=b+1,\dots,D$, the CS condition \eqref{e:comp1} indicates $\alpha_j=0$, and thus
  \begin{align}\label{e:case2}
    1=x_j=y_j-\beta_j+\gamma \le y_j+\gamma,
  \end{align}
  where the last inequality uses the fact that $\beta_j\ge 0$.
\item For $k=a+1,\dots,b$, the CS conditions indicate $\alpha_k=\beta_k=0$, and thus
  \begin{align}\label{e:case3}
    0< x_k=y_k+\gamma <1.
  \end{align}
\end{enumerate}
It is then clear that for any $1\le i \le a$, $b+1 \le j \le D$ and $a+1 \le k \le b$, we have 
\begin{align}
  y_i \le -\gamma < y_k < 1-\gamma \le y_j.
\end{align}
In other words, if the dimensions of $\y$ are sorted in ascending order, the corresponding dimensions of the solution $\x$ is also in ascending order. Therefore, the first step of our algorithm is to sort dimensions of $\y$ into ascending order. And all that is left is to find $(a,b)$, the partition of $\x$ into the three segments. The only KKT condition we have not used so far is the sum constraint \eqref{e:sum}, which now reduces to 
\begin{align}
  \sum_{i=1}^D {x_i}=a\cdot 0 + \sum_{k=a+1}^b (y_k+\gamma) + (n-b)\cdot 1 = s.
\end{align}
This means that if we know $(a,b)$ for the solution $\x$, we must have 
\begin{align} \label{e:gamma}
  \gamma=\frac{s+b-n-\sum_{k=a+1}^b y_k}{b-a}.
\end{align}

Since there are only $\frac{(D+1)(D+2)}{2}$ possible combinations for the indices $(a,b)$, we could test each combination and compute the hypothesized $\gamma$ value using \eqref{e:gamma}. With the hypothesized $\gamma$, the tests we need for $(a,b,\gamma)$ to produce the optimal $\x$ are the following:
\begin{align}
  \label{e:test}
  y_a+\gamma\le 0, \quad\ y_{a+1}+\gamma>0, \quad\ y_b+\gamma<1, \quad\ y_{b+1}+\gamma\ge 1.
\end{align}
It is easy to verify that the $(a,b,\gamma)$ combination that passes the above test leads to $(\x,\balpha,\bbeta,\gamma)$ that satisfy all the KKT conditions, where $\{\alpha_i\}_{i=1}^a$ and $\{\beta_j\}_{j=b+1}^D$ can be retrieved using \eqref{e:case1} and \eqref{e:case2} respectively.
\begin{rmk}
  A special case is when $a=b$, for which \eqref{e:gamma} is ill-defined. In this case, the optimal solution consists of $a$ $0$'s and $n-a$ $1$'s. For this to happen, we must have $s=D-a$ and then \eqref{e:test} reduces to $y_{a+1}+\gamma\ge 1$ and $y_a+\gamma \le 0$, and so $y_{a+1}-y_{a}\ge 1$.
\end{rmk}
\begin{rmk}
  The reason why the computational complexity of our algorithm is $\calO(D^2)$ for \eqref{e:problem} as opposed to $\calO(D\log D)$ for projection onto probability simplex is due to the constraint $\x\le \1$. This constraint is automatically satisfied for the projection onto probability simplex problem where $s=1$, in which case we only need to figure out $a$---the number of zero dimensions in the solution.
\end{rmk}
\begin{rmk} In view of the previous remark, another way of solving \eqref{e:problem} is to alternatively project the estimate onto the (scaled) probability simplex $\{\x:\ \x\ge\0,\ \x^\top \1=s\}$ and the set $\{\x:\ \x \le \1\}$ for which the projection is trivial to compute (we simply threshold the dimensions that are greater than $1$ to $1$). And yet another approach is to apply the Alternating Direction Method of Multipliers \citep{Boyd_11a}, which introduces another copy of the variables $\x$ and alternately optimize each copy with simple steps while encouraging the two copies to agree. We note that these approaches are iterative and the number of iterations depends on the desired accuracy. On the contrary, our method finds the exact solution within a fixed number of steps. 
\end{rmk}

\section{Experiment}

In this section, we demonstrate the effectiveness of our proposed method in Algorithm~\ref{alg1} for solving \eqref{e:problem}. We compare our method with another two solvers. The first one is the CVX package \citep{GrantBoyd12a}, a general convex program solver which transforms the problem into a semi-definite program and then applies interior point method. And the second one is the MATLAB command \texttt{lsqlin} for solving constrained linear least squares. We implement Algorithm~\ref{alg1} in both MATLAB and \texttt{C++} 
and conduct all experiments in MATLAB (the \texttt{C++} code is compiled within MATLAB and the mex-file is used).

All experiments are run on a PC with an Intel Core 2 Quad CPU Q9550 of frequency 2.83GH and 8GB main memory, under Windows 7 and MATLAB version 8.0\@. We generate $\y$ and $s$ using the MATLAB commands 
\begin{equation*}
  \begin{split} 
    &\text{\texttt{y=rand(D,1)-0.5;}} \\
    &\text{\texttt{s=round(rand*D);}}
  \end{split}
\end{equation*}
and record the running time of each method using \texttt{tic} and \texttt{toc}.
We choose the dimension $D$ of $\y$ from $\{50, 100, 500, 1000, 2000, 5000, 10000, 20000, 100000\}$. For each choice of $D$, the experiments are repeated 20 times and the average running times are reported for comparison. 

The results are shown in Table~\ref{tab1}. It can be seen that our proposed method is always faster than CVX and \texttt{lsqlin} for different dimensions $D$ of $\y$. And as $D$ increases, the improvement of our method over the others becomes more significant. If $D$ is relatively large, the compared solvers may run out of memory (denoted as `$-$' in Table~\ref{tab1}). These results confirm that it is beneficiary to explore the special structures of our problem rather than using general convex program solvers. Furthermore, the \texttt{C++} version of our method is by two orders of magnitude faster than the MATLAB version; this improvement is important for projections in very high dimensions.

\begin{table}[t]
  \centering
  \caption{Running time (in seconds) of different solvers for various sizes of $\y$.}
  \label{tab1}
  \begin{tabular}{|c|r|r|r|r|r|r|r|r|r|r|}
    \hline
 \multicolumn{1}{|c|}{Methods} & \multicolumn{1}{c|}{$D=50$}& \multicolumn{1}{c|}{100} & \multicolumn{1}{c|}{500}& \multicolumn{1}{c|}{1000} & \multicolumn{1}{c|}{2000}& \multicolumn{1}{c|}{5000}& \multicolumn{1}{c|}{10000}& \multicolumn{1}{c|}{20000}& \multicolumn{1}{c|}{100000} \\ \hline
    \texttt{lsqlin} & 0.049& 0.11 & 8.18& 59.10 & 360.20 & 5078.00 & -& -& -\\ \hline
    CVX& 0.73 & 1.01 & 5.03& 7.41& 14.55& 40.94& -& -& -\\ \hline
    Ours - MATLAB& 0.0005& 0.002 & 0.023& 0.083& 0.44 & 2.02 & 11.30& 27.09& 870.39 \\ \hline
    Ours - \texttt{C++} & \textbf{0.00002} & \textbf{0.00003} & \textbf{0.0003} & \textbf{0.0009} & \textbf{0.005} & \textbf{0.021} & \textbf{0.11} & \textbf{0.27} & \textbf{8.78} \\ \hline
  \end{tabular}
\end{table}

\bibliographystyle{abbrvnat}

\end{document}

%% file: MACPdef.tex
\newcommand{\x}{\ensuremath{\mathbf{x}}}
\newcommand{\y}{\ensuremath{\mathbf{y}}}

\newcommand{\0}{\ensuremath{\mathbf{0}}}
\newcommand{\1}{\ensuremath{\mathbf{1}}}

\newcommand{\balpha}{\ensuremath{\boldsymbol{\alpha}}}
\newcommand{\bbeta}{\ensuremath{\boldsymbol{\beta}}}

\newcommand{\bbR}{\ensuremath{\mathbb{R}}}

\newcommand{\calL}{\ensuremath{\mathcal{L}}}

\newcommand{\calO}{\ensuremath{\mathcal{O}}}

\newcommand{\norm}[1]{\left\lVert#1\right\rVert}

{%
\begin{list}{#1}{
\vspace{-\topsep}
\vspace{-\partopsep}
\setlength{\itemindent}{0cm}
\setlength{\rightmargin}{0cm}
\setlength{\listparindent}{0cm}
\settowidth{\labelwidth}{#1}
\setlength{\leftmargin}{\labelwidth}
\addtolength{\leftmargin}{\labelsep}
\setlength{\itemsep}{0cm}
}%
}%
{%
\end{list}
\vspace{-\topsep}
\vspace{-\partopsep}
}

{\begin{enumerate}%
}%
{\end{enumerate}}



%% file: MACPdef-ams.tex
\theoremstyle{plain}
\newtheorem{thm}{Theorem}[section]

\newtheorem*{lemma*}{Lemma}

\newtheorem*{prop*}{Proposition}

\theoremstyle{definition}

\newtheorem*{defn*}{Definition}

\newtheorem*{exmp*}{Example}

\newtheorem*{conj*}{Conjecture}

\theoremstyle{remark}
\newtheorem{rmk}[thm]{Remark}
\newtheorem*{rmk*}{Remark}


%% file: note.bbl
\begin{thebibliography}{5}
\providecommand{\natexlab}[1]{#1}
\providecommand{\url}[1]{\texttt{#1}}
\expandafter\ifx\csname urlstyle\endcsname\relax
  \providecommand{\doi}[1]{doi: #1}\else
  \providecommand{\doi}{doi: \begingroup \urlstyle{rm}\Url}\fi

\bibitem[Boyd et~al.(2011)Boyd, Parikh, Chu, Peleato, and Eckstein]{Boyd_11a}
S.~Boyd, N.~Parikh, E.~Chu, B.~Peleato, and J.~Eckstein.
\newblock Distributed optimization and statistical learning via the alternating
  direction method of multipliers.
\newblock \emph{Foundations and Trends in Machine Learning}, 3\penalty0
  (1):\penalty0 1--122, 2011.

\bibitem[Duchi et~al.(2008)Duchi, Shalev-Shwartz, Singer, and
  Chandra]{Duchi_08a}
J.~Duchi, S.~Shalev-Shwartz, Y.~Singer, and T.~Chandra.
\newblock Efficient projections onto the $\ell_1$-ball for learning in high
  dimensions.
\newblock In A.~{McCallum} and S.~Roweis, editors, \emph{Proc. of the 25th Int.
  Conf. Machine Learning (ICML'08)}, pages 272--279, Helsinki, Finland,
  July~5--9 2008.

\bibitem[Grant and Boyd(2012)]{GrantBoyd12a}
M.~Grant and S.~Boyd.
\newblock {CVX}: Matlab software for disciplined convex programming, version
  2.0 beta.
\newblock \url{http://cvxr.com/cvx}, Sept. 2012.

\bibitem[Nocedal and Wright(2006)]{NocedalWright06a}
J.~Nocedal and S.~J. Wright.
\newblock \emph{Numerical Optimization}.
\newblock Springer Series in Operations Research and Financial Engineering.
  Springer-Verlag, New York, second edition, 2006.

\bibitem[Wang and Carreira-Perpi{\~n}{\'a}n(2013)]{WangCarreir13a}
W.~Wang and M.~{\'A}. Carreira-Perpi{\~n}{\'a}n.
\newblock Projection onto the probability simplex: {An} efficient algorithm
  with a simple proof, and an application.
\newblock arXiv:1309.1541, Sept.~3 2013.

\end{thebibliography}
